 \documentclass[sigconf, authorversion]{acmart}
\usepackage{float}
\AtBeginDocument{%
  \providecommand\BibTeX{{%
    \normalfont B\kern-0.5em{\scshape i\kern-0.25em b}\kern-0.8em\TeX}}}

\setcopyright{acmcopyright}
\copyrightyear{2018}
\acmYear{2018}
\acmDOI{XXXXXXX.XXXXXXX}

\acmBooktitle{Woodstock '18: ACM Symposium on Neural Gaze Detection,
 June 03--05, 2018, Woodstock, NY} 
\acmPrice{15.00}
\acmISBN{978-1-4503-XXXX-X/18/06}

\begin{document}

\title{Spatio-temporal Storytelling? Leveraging Generative Models for Semantic Trajectory Analysis}

\author{Shreya Ghosh, Saptarshi Sengupta, Prasenjit Mitra}

\affiliation{%
  \institution{College of IST, Pennsylvania State University}
    \city{State College}
  \state{PA}
  \country{USA}
  \postcode{16801}
}
\email{{shreya, sks6765, pmitra}@psu.edu}

\renewcommand{\shortauthors}{Ghosh and Sengupta, et al.}

\begin{abstract}
  In this paper, we lay out a vision for analysing semantic trajectory traces and generating synthetic semantic trajectory data (SSTs) using  generative language model. Leveraging the advancements in 
  deep learning, as evident by progress in the field of natural language processing (NLP), computer vision, etc.
  we intend to create intelligent models that can study the semantic trajectories in various contexts, predicting future trends,
  increasing machine understanding of movement of animals, humans, goods, etc.
  enhancing human-computer interactions, and contributing to an array of applications ranging from urban-planning to personalized recommendation engines and business strategy.
\end{abstract}




\received{20 February 2007}
\received[revised]{12 March 2009}
\received[accepted]{5 June 2009}

\maketitle

\section{Introduction}
The conventional notion of trajectories, defined as sequences of stop-and-move episodes~\cite{zheng2015trajectory,ghosh2017}, undergoes a transformation when enriched with contextual or semantic information. By augmenting trajectories with semantic tags~\cite{parent2013semantic} that capture additional details such as user activities, environmental conditions, and other relevant aspects, they evolve into spatio-temporal narratives or storytelling. These semantic trajectories weave a comprehensive story that encompasses not only the spatial and temporal dimensions but also the contextual intricacies. This integration of semantics into trajectories enables a deeper understanding and analysis of mobility patterns, paving the way for engaging and coherent spatio-temporal storytelling approaches.

In the field of natural language processing (NLP) and geospatial data, we stand at a remarkable juncture of interdisciplinary exploration~\cite{mai2022towards}. The accelerating pace of technological advancements, combined with the immense potential of deep learning techniques, offers an unprecedented opportunity to redefine the limits of our understanding and capabilities in the context of movement analysis. With the rise of self-attention models, particularly Transformers, in 
NLP, there has been a notable shift from recurrent neural networks (RNNs) to these advanced architectures. Transformers (and transformer-based generative models like GPT) have demonstrated superior performance across diverse NLP tasks, such as text classification, owing to their parallelization capabilities enabling them to ingest data at scales never seen before. Drawing upon this advancement, we propose an approach to analyzing trajectories as narratives, termed as \textit{spatio-temporal storytelling}. By linking semantic trajectories with self-attention models, this article presents a compelling argument for leveraging the power of NLP in the context of spatio-temporal data analysis.
The major objectives of our vision paper are two-folds: (a) we emphasize the need and potential of NLP generative models in the context of semantic trajectory analysis and (b) synthetic semantic trajectory generation using prompt engineering. 
We believe that our proposed vision, although ambitious and challenging, holds the potential for a significant leap in the trajectory data analytics and knowledge extraction domain. For instance, the innovative concept of \emph{Synthetic Semantic Trajectories (SSTs)} using NLP generative models offers potential applications that extend far beyond our current understanding, particularly in fields such as urban planning, transportation, public health, environmental studies, and wildlife conservation. Further, this synthetic data can be used as a proxy for real-life data (such data is sensitive in nature as it captures an agent both in time and location). 

Our proposition of spatio-temporal storytelling and generating SSTs are visionary, aiming to create a new avenue for understanding language (alike) semantics in a spatial and temporal context. This approach can empower decision-makers across various sectors with an unprecedented level of predictive insight, allowing for more informed, proactive, and impactful actions.
\section{Background} 
\subsection{Semantic Trajectory} 
A semantic trajectory~\cite{parent2013semantic} is a modern conceptualization in the field of mobility data analysis, primarily concerned with the integration of semantic information into traditional spatiotemporal trajectories. In its essence, a semantic trajectory is a sequence of chronologically-ordered spatiotemporal points, each associated with a particular semantic meaning that provides context and additional information. This new trajectory dimension moves beyond the mere geographical and temporal aspects to include contextual details such as user activities, environmental conditions, and other aspects of relevance.

Formally, a semantic trajectory $T$ for an object $O$ can be defined as a sequence of tuples:
$T = \{(p_1, t_1, s_1), (p_2, t_2, s_2), \dots , (p_n, t_n, s_n)\}$
where each $p_i$ is a geographical position (a point in a 2-dimensional or 3-dimensional space), each $t_i$ is a timestamp indicating when object $O$ was at position $p_i$, and each $s_i$ is a semantic tag providing additional information about the context at position $p_i$ and timestamp $t_i$. Each semantic tag $s_i$ can be defined as a set of semantic attributes: $s_i = \{a_1, a_2, \dots , a_m\}$, where each $a_i$ represents a distinct type of semantic information. 

While common examples of semantic trajectory include augmenting transportation modes or tourist behaviour, we provide more interesting examples here: 
\begin{itemize}
    \item Semantic trajectories can incorporate \emph{environmental conditions} as semantic tags. For example, consider a trajectory of a delivery vehicle collecting air quality data. Each point in the trajectory includes (lat, lon, time, {pollution level, temperature, humidity}). By analyzing these semantic trajectories, it becomes possible to identify pollution hotspots, study the impact of weather on air quality, and assess the effectiveness of pollution mitigation strategies.
    \item Semantic trajectories can be utilized to \textit{monitor patients' activities and well-being}. For instance, a trajectory could include (lat1, lon1, time1, {home, low activity}), (lat2, lon2, time2, {doctor's office, medical appointment}), (lat3, lon3, time3, {pharmacy, medication purchase}). By examining such trajectories, healthcare providers can gain insights into patients' routines, identify potential health risks, and offer personalized care.
    \item Semantic trajectories can assist in understanding the \textit{impact of events} on urban spaces. Consider a trajectory capturing crowd movement during a music festival. Each point in the trajectory includes (lat, lon, time, {event stage, high density, loud music}). Analyzing these trajectories can reveal patterns of crowd flow, identify areas prone to congestion, and aid in event planning and safety management.
\end{itemize}
The incorporation of contextual information enriches traditional spatiotemporal analysis, enabling deeper insights into mobility patterns, human behavior, and environmental factors.
\subsection{Analysis of Semantic Trajectory}
\begin{enumerate}
    \item Enhanced Narrative Presentation: Spatio-temporal storytelling involves the use of visualizations, narratives, and interactive tools to communicate complex spatio-temporal data in an engaging and understandable manner. By incorporating semantic mobility analysis into spatio-temporal storytelling, the narratives and visualizations can be enriched with contextual information, such as user activities, environmental conditions, or other semantic attributes. \emph{For instance, imagine a study on commuting patterns in a city. A spatio-temporal narrative can be enhanced by integrating semantic data such as traffic conditions, points of interest like coffee shops, or weather conditions. An enriched story would not just describe where and when people move, but also why they choose certain routes (e.g., due to traffic congestion, proximity to a popular cafe, or better weather conditions).}

\item Contextual Insights: Semantic mobility analysis aims to integrate semantic information into traditional spatio-temporal trajectories, providing context and additional information beyond geographical and temporal aspects. Spatio-temporal storytelling can leverage these contextual insights to enhance the storytelling process. By incorporating semantic tags and attributes, the narratives can provide richer descriptions of the places visited, activities performed, or the impact of environmental factors on mobility patterns. This adds depth and relevance to the storytelling by incorporating meaningful context. \emph{For example,  in tracking the movement of a predator species, semantic data about prey movements, terrain characteristics, or seasonal changes can provide meaningful context. The storytelling can then reveal insights about the predator's hunting strategies, preferred habitats, or seasonal adaptations.}

\item Exploratory Analysis: Spatio-temporal storytelling often involves interactive exploration of data, allowing users to navigate through different time periods, locations, or specific events. By integrating semantic mobility analysis, users can delve deeper into the data during the storytelling process. They can interactively explore trajectories with specific semantic attributes, filter trajectories based on activities or environmental conditions, or focus on specific subgroups within the data. This allows for a more interactive and personalized storytelling experience. \emph{For instance, in a study on elephant movements, users could interactively explore the data based on semantic attributes like proximity to water sources, human settlements, or vegetation density. This could reveal patterns on how these factors influence the elephants' movement and habitat choices.}

\end{enumerate}
\subsection*{Why Generative Language Models?}
While several deep learning models including RNNs (Recurrent Neural Networks), LSTMs (Long Short-Term Memory), and GANs (Generative Adversarial Networks)~\cite{su2020survey,henke2023condtraj} have been used effectively in various time-series applications, including mobility analysis, however, they have certain limitations that make Generative Language Models (GLMs) with prompt engineering more advantageous in the context of semantic mobility trajectory analysis. 
While GANs are great for generating new samples that resemble the training data, they often lack control over what specifically is generated. Similarly, controlling the outputs of LSTMs or RNNs can be challenging. In contrast, with GLMs, one can use prompt engineering to guide the model's output, providing a degree of control over the generated response. This is especially useful when we want to extract specific types of information or patterns from the data. Also, While LSTMs and RNNs can handle sequences, they often struggle with long-range dependencies due to the vanishing gradient problem. On the other hand, GLMs like the transformer-based GPT-4 use a mechanism called self-attention, which allows them to model dependencies between any two points in the sequence, regardless of their distance. This capability is crucial when dealing with complex mobility trajectories that span long timeframes. Finally, Transformer-based models like GPT-4 are more parallelizable compared to RNNs and LSTMs, which need to process sequences step-by-step. This allows GLMs to scale better with large datasets, which is often a requirement in mobility analysis due to the large amounts of data generated by GPS devices, social media, etc.
\section{Synthetic Semantic Trajectory Generation}
\emph{Context-Aware Synthetic Trajectories:}
Using prompt engineering, we can develop prompts that capture contextual information such as time of day, weather conditions, user preferences, and environmental factors. By conditioning the generation process on these prompts, NLP models can generate synthetic trajectories with realistic semantic tags that reflect the given context. For example, a prompt like ``Generate a semantic trajectory of a tourist in a coastal city during a sunny day" can produce trajectories with tags related to beach activities, sightseeing, and outdoor dining.

\emph{Personalized Trajectory Generation:}
Create prompts that incorporate user profiles and preferences to generate personalized semantic trajectories. By considering factors like age, interests, and mobility patterns, NLP models can generate synthetic trajectories tailored to specific individuals. For instance, a prompt like ``Generate a semantic trajectory for an adventurous young traveler interested in hiking and photography" can produce trajectories with tags related to hiking trails, scenic viewpoints, and photography spots.

\emph{Transfer Learning for Semantic Trajectories:}
We can leverage transfer learning techniques in prompt engineering to generate semantic trajectories in different domains or locations. First, we train NLP models on existing semantic trajectory datasets and fine-tune them using prompts specific to a new domain or location. This allows for the generation of synthetic trajectories that capture the semantic characteristics and patterns observed in the target domain or location, enabling scenario-specific analysis and planning.

\emph{Multimodal Semantic Trajectories:}
We can incorporate multimodal prompts that encompass multiple sources of semantic information, such as textual descriptions, images, or audio data. By leveraging multimodal data in prompt engineering, NLP models can generate synthetic trajectories with diverse semantic tags that reflect the fusion of different modalities. For example, prompts that combine textual descriptions of locations with corresponding images (visual prompts) can generate trajectories with tags capturing both visual and textual context.

The problem can be formulated as follows:

Given a set of observed mobility trajectories $T = \{T_1, T_2, \ldots, T_n\}$, where each trajectory $T_i = \{(p_1, t_1, s_1), (p_2, t_2, s_2), \ldots, (p_m, t_m, s_m)\}$ is a sequence of $m$ tuples. Each tuple contains a spatial location ($p$), a timestamp ($t$), and a semantic label ($s$) indicating the activity or context associated with the location at each point. Our goal is to utilize a generative model $G$ that, given a trajectory sequence $T_i$, can generate a synthetic trajectory sequence $T'_i = G(T_i)$. We represent each location $p$ in a trajectory by a spatial coordinate vector, $p = (x, y)$, each timestamp $t$ in a standard format, such as Unix time, and the semantic label $s$ as a vector in a semantic space, $s = (s_1, s_2, \ldots, s_k)$, where $k$ is the number of distinct semantic labels.
The generative model $G$ needs to learn the function $f$, mapping the input sequence $T_i$ to an output sequence $T'_i$: $G(T_i) = T'_i = f(T_i)$, where $f$ is the function representing the patterns and structures inherent in the trajectory data, including the spatial, temporal, and semantic dimensions. To train the model, we minimize a loss function $L$ which measures the discrepancy between the synthetic trajectories and the observed trajectories: $\min_G L(T_i, G(T_i)) = \min_G L(T_i, T'_i)$. 
The generated synthetic trajectories should maintain the integrity of the original data, capturing essential characteristics such as spatial distribution, temporal trends, and sequence patterns. Simultaneously, the model should ensure the diversity and novelty of the generated trajectories, enabling them to be used for various applications, including simulation, prediction, or data augmentation for real-world datasets.

\section{Generative Models and Prompt Engineering for Semantic Mobility Trajectories}
\subsection*{Generative Models}

Generative models are a class of statistical models that allow for the generation of new instances of data, based on patterns learned from a given dataset. In the context of synthetic semantic mobility trajectories, we focus on generative models inspired by NLP techniques 
owing to them capturing sequential and semantic patterns.

\noindent \textbf{Autoregressive Models:} Models such as GPT (Generative Pre-trained Transformer) learn to generate new sequences based on the preceding context. They could be adapted to our problem by treating location-time-semantic tuples as tokens and learning to generate subsequent tuples based on preceding ones, akin to \cite{bosselut-etal-2019-comet} who trained a GPT-checkpoint on knowledge-graph (KG) (subject, predicate) pairs with the goal of predicting the object of the tuple for the task of KG completion.


\subsection*{Prompt Engineering}

Prompt engineering is a crucial aspect of applying generative models, particularly transformer-based models like GPT, to specific tasks. A prompt specifies the task the model is to perform and guides its generation process. In our case, prompt engineering involves designing an effective way to present the trajectory data to the model and to specify the task of generating a synthetic trajectory. For example, we could format each trajectory sequence as a text sentence, with each location-time-semantic tuple represented as a word, and the task of generating a synthetic trajectory framed as a sentence completion task.
This can be formulated mathematically as follows:
Given a trajectory $T = (p_1, t_1, s_1), \ldots, (p_m, t_m, s_m)$, we represent it as a sentence $S = w_1, \ldots, w_m$, where each word $w_i$ corresponds to a tuple $(p_i, t_i, s_i)$. The task of generating a synthetic trajectory is then framed as the task of generating a sentence $S' = w'_1, \ldots, w'_{m'}$, where each $w'_i$ corresponds to a synthetic tuple $(p'_i, t'_i, s'_i)$.
The quality of the prompts can significantly affect the model's performance, making prompt engineering a vital part of the model design process. 

\subsection{Prompt Engineering for Semantic Mobility Trajectories}
Traditionally, trajectory sequences are viewed as a mere chronological ordering of locations, time-stamps, and associated semantic labels. In our novel approach, we propose treating a trajectory as a \textit{"story"} of an entity's movement through space and time.

\subsubsection{Narrative Encoding of Trajectories}

In this approach, each trajectory $T = (p_1, t_1, s_1), \ldots, (p_m, t_m, s_m)$ is converted into a narrative $N = w_1, \ldots, w_m$, where each $w_i$ represents a \textit{"sentence"} that describes an entity's state, based on the corresponding tuple $(p_i, t_i, s_i)$. The sentences should be constructed to clearly express the spatial location, time, and semantic context, and also to create a narrative flow that mirrors the entity's movement.

For example, a tuple representing a person being at a coffee shop at 10:00 AM might be represented as: "At 10:00 AM, the person was enjoying a hot coffee at the bustling cafe located at coordinates (x, y)." The subsequent tuple might indicate a change in location and activity, represented as a new sentence, thus continuing the narrative. Models such as Galactica \cite{taylor2022galactica} would be well suited for this task owing to it being trained explicitly on reasoning problems and scientific text. The choice of model is critical here since we need them to reason about mathematical properties of data rather than generating unwanted output as multilingual text (BLOOM \cite{scao2022bloom}).

\subsubsection{Story Completion as Trajectory Generation}

The task of generating a synthetic trajectory is then framed as a story completion task. Given the narrative $N = w_1, \ldots, w_k$ representing the initial part of a trajectory, the model is tasked with generating the remainder of the story, $N' = w'_{k+1}, \ldots, w'_{m'}$, which corresponds to the remainder of the synthetic trajectory.

This approach leverages the ability of modern NLP models to understand and generate narrative structures, potentially enabling them to better capture the patterns and dependencies in the trajectory data. It also opens up the possibility to incorporate more complex semantic information into the trajectories, such as the entity's goals or preferences, making the synthetic trajectories richer and more realistic.
\subsubsection{Visual Prompting} 
\begin{figure}[ht]
    \centering
    \includegraphics[width=0.45\textwidth]{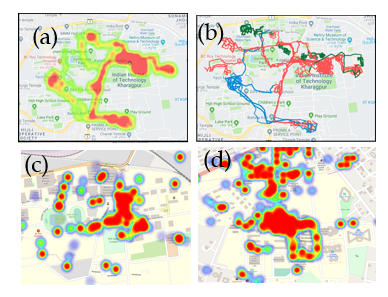}
    \caption{Trajectory Visualization over IIT Kharagpur and NIT Warangal Campus (a) Heatmap of user's daily movement; (b) Cluster of similar movement behaviours depicted in three different colors; (c) Heatmap of aggregated movement on IIT Kharagpur; (d) Heatmap of aggregated movement on NIT Warangal }
    \label{fig:example}
\end{figure}

Text-to-image models, like DALL-E\footnote{\url{https://en.wikipedia.org/wiki/DALL-E}}, or Stable Diffusion, represent a significant advancement in the capabilities of generative models, enabling the transformation of textual prompts into detailed and contextually relevant images. For instance, if we have textual data about the movements or trajectories within academic campuses, text-to-image models could generate images that visually depict the described scenarios. For instance, given a prompt like "Visualize the common route students take from the dormitory to the main library in the IIT Kharagpur campus", a model like DALL-E could generate an image that provides a visual interpretation of this route, which can then be compared with the existing cluster or heatmap images for validation or further analysis (See Fig.~\ref{fig:example}).

Visual prompting can offer several benefits in semantic trajectory analysis as: (a) Visual Interpretation of Trajectories: Visual prompts can enable the generation of graphical representations of semantic trajectories, providing a more intuitive way to understand and interpret mobility patterns. For instance, a textual prompt such as ``Generate a visual semantic trajectory for a daily commute in IIT Kharagpur campus during rush hour" could produce an image that depicts the spatial-temporal path along with associated traffic conditions, points of interest, or other relevant factors. (b) Contextual Visualization: Visual prompts could be used to generate images that reflect the context associated with specific parts of a trajectory. For example, a prompt describing a walk through a park might generate an image showing the park setting, while a subsequent prompt describing a visit to a nearby café might produce an image of the café. These contextually relevant images can enrich the semantic understanding of the trajectory. (c) Data Augmentation: Text-to-image models can be used to create synthetic images based on generated trajectories, effectively augmenting the dataset. These synthetic images can be useful in training or testing other models, such as computer vision models that analyze images to understand mobility patterns. On the other side, if we want to generate descriptive narratives from the given images, models like CLIP from OpenAI, which understands both text and images, could be helpful. We can feed the image (heatmaps) to such a model with a text prompt like ``Describe the pattern of movement in this image", and the model could generate a textual description that enriches our understanding of the image. (d) Multi-Modal Analysis: The integration of visual prompting can enhance multi-modal analysis, incorporating visual context into the understanding and generation of semantic trajectories. For example, a visual prompt combined with a text description can generate a trajectory that captures both visual and textual semantics, offering a richer and more comprehensive depiction of the scenario. 
\subsubsection{Evaluation of Prompts}

The success of this novel prompting strategy, like any other, ultimately needs to be evaluated empirically. Different narrative structures and levels of detail in the sentence representations should be experimented with, and their impact on the quality of the generated trajectories assessed. Evaluation metrics should consider not only the spatial-temporal accuracy of the synthetic trajectories but also their semantic coherence and the narrative flow they create.

\subsection{Chain of Thoughts: Approach for Semantic Mobility Trajectories}

The Chain of Thoughts approach represents an innovative leap in prompt engineering for semantic mobility trajectory generation. This method stems from the concept that each individual's movement is a manifestation of their underlying thought process. By encoding this "chain of thoughts" into our prompt engineering, we can model an entity's trajectory as a narrative that reflects their decision-making process, rather than just a series of location-time-activity events.

\subsubsection{Thought-Encoded Trajectories}

In the Chain of Thoughts approach, each trajectory $T = (p_1, t_1, s_1), \ldots, (p_m, t_m, s_m)$ is encoded into a narrative $N = w_1, \ldots, w_m$, where each $w_i$ is a \textit{"thought sentence"} that describes not only the location, time, and semantic activity represented by the corresponding tuple $(p_i, t_i, s_i)$, but also the thought process leading to the movement.

For example, a tuple representing a person at a coffee shop at 10:00 AM might be represented as: "Feeling the need for a caffeine boost at 10:00 AM, the person decided to visit the nearby coffee shop located at coordinates (x, y)." Here, the sentence doesn't merely state the location and activity, but also incorporates the hypothetical thought process leading to that action.

\subsubsection{Thought Generation as Trajectory Generation}

The task of generating a synthetic trajectory is then framed as a "thought generation" task. Given the thought-encoded narrative $N = w_1, \ldots, w_k$ of the initial part of a trajectory, the model is tasked with generating the remainder of the chain of thoughts $N' = w'_{k+1}, \ldots, w'_{m'}$ or even track back and generate the thought process leading up to that event, as mentioned above. 

This method leverages the capability of modern NLP models to understand and generate coherent narratives, thereby potentially capturing more nuanced patterns in trajectory data. Furthermore, this approach enables the inclusion of more complex semantic information into the trajectories, such as motivations, preferences, or environmental factors influencing an entity's movements, making the synthetic trajectories richer, more realistic, and more interpretable.



\subsection{Time-Geography Inspired Prompt Engineering and Generative Modelling}

Time-geography~\cite{miller2005measurement} is a classic approach to understanding human mobility, which considers the constraints of time and space in shaping the trajectories of individuals. This perspective offers a unique opportunity for novel prompt engineering and the design of generative models for synthetic semantic mobility trajectory generation.

\subsubsection{Time-Space Prisms in Prompt Engineering}

In the time-geography framework, a key concept is the Time-Space Prism~\cite{yu2019direction}, which represents the set of all points that an individual could theoretically reach, given the constraints of time and their travel speed. These prisms provide a geometric representation of the potential paths that an individual can take.

In our framework, we propose encoding these Time-Space Prisms into our prompts, in addition to the actual trajectory data. Each prism could be represented as a multi-dimensional interval that encapsulates the range of possible locations and times that an individual can reach, based on their current location, time, and travel speed. This added information gives the generative model more context about the possible future paths of an individual, beyond what is provided by their past trajectory alone. This could potentially improve the realism of the synthetic trajectories generated by the model, by ensuring they are consistent with the individual's time-space constraints.

\subsubsection{Generative Model Design}

The generative model design would need to be adapted to handle the additional time-space prism data. A possible approach would be to use a variant of the Transformer model for trajectory classification~\cite{liang2022trajformer}, which can handle multi-dimensional input data and learn complex dependencies between them.

The model would be trained to generate synthetic trajectories that are not only similar to the observed trajectories but also consistent with the associated time-space prisms. This would involve defining a suitable loss function that encourages the generated trajectories to stay within the boundaries defined by the time-space prisms.
\subsection{Reinforcement learning based Prompt engineering}
Reinforcement Learning (RL) combined with prompt engineering is particularly useful for semantic trajectory analysis due to the specific nature and demands of this task: \emph{High Dimensionality:} Semantic trajectory analysis often involves dealing with high-dimensional data, which includes not only spatial-temporal information but also diverse semantic attributes. RL with prompt engineering can help to navigate this complex space more effectively by learning to select prompts that generate meaningful trajectories. \emph{Dynamic Contexts:} Real-world mobility contexts are highly dynamic, with constantly changing environmental factors, user behaviors, and other variables. RL can adapt to these changes over time, iteratively learning from feedback to improve the generation of contextually relevant trajectories. \emph{Goal-oriented Outputs:} In semantic trajectory analysis, there is often a specific goal or criteria for the generated trajectories, such as realism, relevance to a particular user profile, or adherence to certain environmental constraints. RL provides a framework for optimizing these goal-oriented outputs, by incorporating feedback on the generated trajectories into the learning process. On the contrary, basic language understanding can often be achieved without RL. General-purpose language models like GPT-4 are trained on a broad range of internet text and learn to generate coherent and contextually appropriate responses by predicting the next word in a sequence. They can understand and generate language based on this unsupervised learning process, without needing explicit reward signals or carefully engineered prompts.

Consider the task of generating semantic mobility trajectories using a language model. We denote the set of all possible prompts as $P$, and the prompt selected in each step as $p_t \in P$, where $t$ is the time step. The language model generates a trajectory $T = G(p_t)$ in response to prompt $p_t$, where $G$ represents the generative function of the language model. The trajectory $T$ is then evaluated by a reward function $R(T)$ which assigns a score to quantify the quality of the generated trajectory.

Our goal is to learn a policy $\pi$ that maximizes the expected reward over a sequence of $N$ steps. The policy $\pi$ is a distribution over $P$ that dictates the probability of choosing each prompt. We can formalize this as an optimization problem:
\begin{equation}
\max_{\pi} \; E_{p_t \sim \pi} \left[ \sum_{t=1}^{N} R(G(p_t)) \right]
\end{equation}
We use a reinforcement learning approach to solve this problem. In each step, an action (i.e., prompt) is selected according to policy $\pi$, a trajectory is generated and evaluated to obtain a reward, and the policy is updated based on the observed reward. 
A common method to update the policy in RL is the policy gradient method. For a policy parameterized by $\theta$, the update rule in policy gradient method can be written as:
\begin{equation}
\Delta\theta = \alpha \nabla_\theta \log \pi(p_t|\theta) R(G(p_t))
\end{equation}
Where:
- $\alpha$ is the learning rate.
- $\nabla_\theta \log \pi(p_t|\theta)$ is the gradient of the log-probability of the chosen prompt $p_t$ with respect to the policy parameters $\theta$.
- $R(G(p_t))$ is the reward of the trajectory generated using prompt $p_t$.

This process is repeated over multiple episodes, allowing the model to explore different prompts, generate a variety of trajectories, and incrementally improve the policy to favor prompts that yield higher rewards. To enhance exploration of the prompt space, strategies like $\epsilon$-greedy or entropy regularization can be employed, adding an additional term to the reward function that encourages diversity in prompt selection. This RL-based approach to prompt engineering enables the generation of high-quality semantic mobility trajectories that are tailored to specific evaluation criteria, adapt to dynamic contexts, and effectively navigate the high-dimensional space of possible trajectories.

\emph{Running example:} Let's consider a use case of planning an itinerary for a tourist visit in a city like New York using a language model. The goal here is to generate a semantic trajectory that describes an optimal itinerary based on the tourist's preferences (such as visiting landmarks, tasting local cuisines, enjoying the nightlife, etc.) and other contextual factors (such as time of day, weather conditions, local events, etc.). In this scenario, the set of prompts $P$ might include prompts like: (1) "Generate a morning itinerary that includes landmarks and local breakfast spots in New York." (2) "Create an afternoon itinerary with art galleries and a local dining experience in New York."
"Plan an evening itinerary with Broadway shows and nightlife in New York." (3) For each prompt $p_t$, the language model generates a trajectory $T = G(p_t)$, which describes a possible itinerary. The reward function $R(T)$ could be based on criteria such as: (a) The relevance of the itinerary to the given prompt. (b) The diversity of activities in the itinerary. (c) The feasibility of the itinerary considering travel distances and durations.
User feedback on the proposed itinerary.
To start with, the policy $\pi$ might assign equal probabilities to all prompts. But as the reinforcement learning process proceeds, it would learn to favor prompts that yield higher rewards. For example, it might learn that morning itineraries that include famous landmarks like the Statue of Liberty followed by a visit to a popular local breakfast spot generate higher rewards. Similarly, an afternoon prompt focusing on visiting art galleries and unique dining experiences might be favored. The policy might also adapt to different contexts. For instance, on a rainy day, the policy might give higher probability to prompts that involve indoor activities. With the reinforcement learning approach, the model can effectively learn to generate contextually relevant and user-preferred itineraries by exploring and exploiting the prompt space.

Apart from synthetic trip generation, the method can be used for varied semantic mobility analytics tasks: (1) Optimal Route Recommendation: Leveraging RL, language models can be used to recommend the most optimal route given certain conditions. The reward can be defined in terms of the shortest distance, least time, or maximum sightseeing spots. The model can suggest different optimal routes by dynamically choosing prompts conditioned on the context (weather, traffic conditions, time of day, etc.). (2) Predicting Future Mobility Patterns: By training language models using historical semantic trajectories and using RL for prompt engineering, we could generate prompts that lead to the prediction of future mobility patterns. The model could be rewarded for accurately predicting these patterns based on historical data. (3) Semantic Hotspot Identification: Prompts could be generated to identify popular locations or activities in a certain geographic area over a period. These hotspots could then be used for urban planning, traffic management, or even for business purposes such as targeted advertising. (4) Anomaly Detection in Mobility Patterns: Prompts can be engineered to identify unusual patterns or anomalies in mobility data. RL can be used to guide the model towards identifying these anomalies accurately. The model could be rewarded for correctly identifying anomalous patterns while penalized for false positives or negatives. (5) Understanding Mobility Behavior: By generating prompts that aim to capture user activity, preferences, or behavior in mobility data, RL can be used to understand and model mobility behavior better. For example, prompts could be generated to understand the preferred mode of transport, frequently visited locations, or activity patterns of different user groups. The key idea here is to learn a policy for selecting prompts that maximize the reward, which could be any measure of success depending on the specific task at hand.




\subsection{Semantic Interpretation of Mobility Patterns}

Generative NLP models, especially transformer-based models like GPT, have been trained on vast amounts of text data and have shown impressive abilities in understanding and generating human-like text. When applied to mobility analysis, these models can generate human-readable narratives that represent mobility trajectories. By transforming raw location-time-activity tuples into semantically rich narratives, these models can provide a more intuitive and contextually grounded interpretation of mobility patterns.

\subsubsection*{Explainability through "Chain of Thoughts" Approach}

As previously described, the Chain of Thoughts approach involves encoding the hypothetical thought processes that lead to each movement event in a mobility trajectory. By generating narratives that not only describe the movements but also the motivations and decisions behind them, generative NLP models can provide explanations for why certain mobility patterns occur. This could make the models' outputs more understandable and useful for human users.

\subsubsection*{Explainability through Time-Space Prism Constraints}

Incorporating time-space prism constraints into the prompts provided to the model can also enhance explainability. By generating trajectories that respect these constraints, the model demonstrates an understanding of the physical and temporal limitations that govern real-world mobility. Comparing the synthetic trajectories with the time-space prisms can provide insights into how the model navigates the available paths and which factors it considers when making decisions. This can offer a geometric and temporal perspective on the model's decision-making process, further enhancing its explainability.



\subsubsection*{Exploration of Counterfactual Scenarios}

Finally, NLP generative models can also help explore counterfactual scenarios, which is a powerful tool for understanding the causal relationships in data. By modifying certain aspects of the prompts (e.g., changing the time or location of certain events) and observing how the model alters the generated trajectories in response, we can infer how these changes might affect real-world mobility patterns. This can provide a form of model-based reasoning that can complement and enrich traditional data-driven mobility analyses.

\subsubsection*{Semantic Trajectory Summarization}

Much like how NLP models can summarize lengthy articles or reports into a few sentences, we can develop models to summarize complex trajectories into a few key semantic points or activities. This can facilitate understanding the key points of a trajectory without delving into intricate details, and could be especially useful in large-scale studies of human mobility where individual trajectory details might be less important than overall patterns.

\subsubsection*{Trajectory Style Transfer}
Inspired by the concept of style transfer in text and images, we consider a similar concept for trajectories. For example, given the trajectory of a particular individual or vehicle, we aim to generate a new trajectory that maintains the same key locations or activities but alters the style of movement (e.g., changing a car's trajectory into a biking trajectory while visiting the same locations).
NLP generative models can generate counterfactual narratives based on given scenarios. Similarly, given a certain trajectory, these models could generate counterfactual trajectories under different hypothetical scenarios, helping us understand how changes in conditions might affect mobility patterns.

\subsubsection*{Interactive Trajectory Generation}

Taking inspiration from interactive story generation in NLP, we can envision a system where users interactively guide the generation of a trajectory, providing input or making choices at certain points, and the model generates the remainder of the trajectory accordingly.

\subsubsection*{Contextual Anomaly Detection}

By training NLP generative models to understand normal patterns in trajectory data, they could be used to detect anomalies or unusual events. Unlike traditional anomaly detection methods, these models could consider the semantic context of the trajectories, potentially detecting more subtle or complex anomalies.

\subsubsection*{Synthetic Environment Generation}

NLP models can generate not only synthetic trajectories, but also synthetic environments within which these trajectories exist. This could involve generating descriptions of locations or activities at those locations, and could be used to create immersive virtual worlds for simulations or games.

\subsubsection*{Cross-modal Trajectory Analysis}

By treating trajectories as text-like sequences, we could use NLP models to analyze trajectories alongside other types of sequential data, such as text or time-series data. This could enable richer, cross-modal analyses of human behavior, such as correlating mobility patterns with social media activity or economic trends.

In all these applications, NLP generative models can leverage their powerful sequence understanding and generation capabilities to provide novel perspectives and methods for spatio-temporal data mining.

\section{Conclusion}
In summary, the paper presents a comprehensive framework for semantic trajectory analysis and generation, leveraging prompt engineering and generative language models. The proposed approaches offer promising directions for generating trajectories that capture contextual information, personalize to specific individuals, transfer knowledge across domains, and incorporate multimodal data. We discuss the application of time-geography principles in prompt engineering, incorporating time-space prisms into prompts to provide more context and improve the realism of generated trajectories. Furthermore, we propose the use of reinforcement learning combined with prompt engineering to navigate the high-dimensional and dynamic nature of semantic trajectory analysis and optimize goal-oriented outputs. These advancements contribute to scenario-specific analysis, planning, and various applications, including simulation, prediction, and data augmentation for real-world datasets.

\end{document}